%% file: main.tex
\documentclass[runningheads]{llncs}

\usepackage{eccv}

\usepackage{eccvabbrv}

\usepackage{graphicx}
\usepackage{booktabs}
\usepackage{multirow}
\usepackage{tcolorbox}
\usepackage{xcolor}
\usepackage{caption}
\definecolor{varcolor}{RGB}{0, 0, 255}
\usepackage[accsupp]{axessibility} 
\usepackage{hyperref}
\usepackage{xurl}

\begin{document}

\title{EgoSelf: From Memory to Personalized Egocentric Assistant}

\author{Yanshuo Wang$^{*}$\inst{1,2} \and
  Yuan Xu$^{*}$\inst{3} \and
  Xuesong Li\inst{4} \and
  Jie Hong\inst{5} \and
  Yizhou Wang\inst{3} \and
  \\
  Chang Wen Chen$^{\dagger}$\inst{2} \and
  Wentao Zhu$^{\dagger}$\inst{1}}

  \authorrunning{Y.~Wang et al.}

\institute{Eastern Institute of Technology, Ningbo \and
Hong Kong Polytechnic University \and
Peking University \and
Commonwealth Scientific and Industrial Research Organization \and University of Hong Kong}
\maketitle

\begin{abstract}
Egocentric assistants often rely on first-person view data to capture user behavior and context for personalized services. Since different users exhibit distinct habits, preferences, and routines, such personalization is essential for truly effective assistance. However, effectively integrating long-term user data for personalization remains a key challenge. To address this, we introduce EgoSelf, a system that includes a graph-based interaction memory constructed from past observations and a dedicated learning task for personalization. The memory captures temporal and semantic relationships among interaction events and entities, from which user-specific profiles are derived. The personalized learning task is formulated as a prediction problem where the model predicts possible future interactions from individual user's historical behavior recorded in the graph. Extensive experiments demonstrate the effectiveness of EgoSelf as a personalized egocentric assistant. Code is available at \url{https://abie-e.github.io/EgoSelf/}.
\end{abstract}

\input{sections/1_introduction}

\input{sections/2_related_work}

\input{sections/3_method}
\input{sections/4_experiments}

\input{sections/5_conclusion}

\bibliographystyle{splncs04}
\bibliography{main}

\end{document}

%% file: sections/1_introduction.tex
\section{Introduction}
Egocentric data captures how we naturally perceive and interact with the surrounding world, providing rich contextual information while implicitly encoding individual behavioral patterns and preferences. With the rapid advancement of AR/VR glasses, capturing such first-person data has become increasingly accessible, and effectively understanding this continuous stream of personal experience can facilitate a wide range of personal tasks~\cite{mangalam2023egoschema, yang2025egolife, chen2023egoplan, grauman2024ego}.
Consider a life assistant integrated with AR glasses, as shown in Figure~\ref{fig:front}, that can remind you where you last placed your keys, or suggest preferred dishes at a restaurant. Such support is inherently user-specific: different individuals follow distinct daily routines, hold different preferences, and develop unique interaction habits, making personalization tailoring responses to each user's history and behavior indispensable for a truly effective assistant. Achieving this, however, presents two intertwined challenges: \textit{(i)} effectively organizing the continuous, unstructured egocentric stream into a structured memory that supports efficient retrieval, and \textit{(ii)} moving beyond factual recall to internalize user-specific behavioral patterns for genuine personalization.

\begin{figure}[t]\centering
\includegraphics[width=1.0\linewidth]{figure/teaser_submit.pdf}
\vspace{-20pt}
\caption{An example: Egocentric personal assistants capture multi-day egocentric activity history to construct structured interaction memory. From this memory, they summarize general events and temporal patterns to build a persistent user model that encodes personal preferences and daily habits. Finally, they generate consistent and personalized responses that are well aligned with the user’s inherent behavior patterns and situational context.}
\vspace{-16pt}
\label{fig:front}
\end{figure}

Egocentric video constitutes a continuous, extremely long, and unstructured multimodal stream, in which meaningful information is buried amid vast amounts of redundant content. With recent advances in Multimodal Large Language Models (MLLMs)~\cite{chenlongvila, cheng2024egothink, zhang2024long, qu2025does}, long-context vision-language models~\cite{zhang2024long,shenlongvu, sun2019videobert, lin2024vila, weng2024longvlm} can process extended videos through feature compression or context extension, yet they treat all content uniformly without structured modeling of interactions or their inter-relations. Work in egocentric vision~\cite{peirone2024backpack,lin2022egocentric, li2021ego, moon2022imu2clip, pramanick2023egovlpv2,wang2023learning,wang2023ego} has advanced first-person action recognition, but has primarily focused on short clips and general activity understanding rather than on long-term behavioral organization. Even retrieval-augmented approaches face severe precision degradation at scale. For instance, EgoRAG~\cite{yang2025egolife} retrieves relevant episodes based on embedding similarity, but as the user history grows, semantically similar yet contextually irrelevant entries increasingly overwhelm simple similarity-based matching.

Yet even when past experiences are effectively organized, a fundamental gap remains between remembering what happened and understanding the user. Locating a misplaced item requires only factual retrieval, but anticipating a user's preferences or predicting their likely next actions demands internalized knowledge of their behavioral regularities and latent tendencies. While some egocentric work~\cite{chen2023egoplan, perrett2025hd, damen2018scaling} captures detailed human-object interactions, it focuses only on short video segments and does not model long-term user-specific patterns. To date, no existing framework simultaneously addresses the structured organization of long-term multimodal egocentric data and the internalization of user-specific behavioral patterns for genuine personalization.

To address this gap, we draw inspiration from Conway's Self-Memory System (SMS) \cite{conway2000construction} and propose EgoSelf, a framework that progressively builds personalized understanding from first-person experiences. The SMS framework emphasizes a hierarchical, interconnected memory system composed of two complementary components: episodic memory that records concrete, context-rich personal events and experiences, and semantic self-memory that abstracts and stores stable, generalized knowledge about the self over time. Most importantly, it posits that human experiences are not stored as isolated episodes but organized into hierarchical structures, with specific events progressively linked to broader patterns and themes. Individuals construct their self-knowledge upon this foundation, developing a stable understanding of their own habits, preferences, and behavioral tendencies \cite{conway2005memory}.

Following this principle, EgoSelf segments continuous egocentric streams into discrete interaction events, each of which captures a moment when the user actively engages with objects or people. It then extracts the objects and persons involved in these events and consolidates them into persistent entity nodes that bridge related episodes over time. Temporal, causal, and co-activity edges between events, together with event-entity associations, form a heterogeneous interaction graph that transforms fragmented observations into a structured, queryable memory supporting efficient retrieval and pattern-level reasoning. A compact user profile is further derived from the graph by clustering recurring behaviors and summarizing long-term preferences.

Building upon this organized memory, EgoSelf internalizes user-specific behavioral patterns into model parameters. Motivated by the cognitive insight that genuine understanding is reflected in the ability to anticipate~\cite{bruner2009process, wang2025fosteringvideoreasoningnextevent}, we formulate a self-supervised habit-learning task in which the model is trained to predict a user's subsequent interactions, given their historical interaction chain in the graph. This prediction objective drives the model to capture individual behavioral regularities and latent preferences. At inference time, the graph supports efficient retrieval through entity-anchored filtering and relation-guided context expansion, and the model generates personalized responses grounded in the retrieved context, the user profile, and the learned behavioral patterns. In summary, our main contributions include:
\begin{itemize}
    \item We propose EgoSelf, a framework for personalized egocentric assistants that organizes long-term first-person observations into a heterogeneous interaction graph with persistent entity nodes and compact user profiles.
    \item To internalize user-specific behavioral patterns for personalization, we design a self-supervised habit learning task that predicts future interactions from historical graph data.
    \item Extensive experiments on the EgoLifeQA benchmark demonstrate that EgoSelf achieves state-of-the-art performance across diverse personalization tasks.
\end{itemize}

%% file: sections/2_related_work.tex
\section{Related Work}
\subsection{Long-Context Video Language Models}
Processing long-form videos poses significant challenges for vision-language models (VLMs) due to the large number of visual tokens required~\cite{wang2025fostering, zhang2024long, chenlongvila}. Recent efforts address this through two main strategies: token compression and memory augmentation. Compression-based methods reduce visual redundancy to fit extended footage within context limits: LongVU~\cite{shenlongvu} employs spatiotemporal adaptive compression using DINOv2 features to identify and remove redundant frames, enabling hour-long video processing within 8K context windows; LLaMAVID~\cite{li2023llamavid} achieves extreme compression by representing each frame with only two tokens; and SlowFast-LLaVA~\cite{xu2024slowfast} adopts a two-stream design that balances spatial detail extraction with temporal motion capture without additional training overhead. Memory-augmented approaches maintain external storage for historical information: MA-LMM~\cite{he2024malmm} introduces a memory bank that stores past video features for online long-term analysis, MovieChat~\cite{song2024moviechat} first draws on the Atkinson-Shiffrin model with dual short- and long-term memory to handle over 10K frames, and VideoStreaming~\cite{qian2024videostreaming} propagates memory across segments with adaptive retrieval for question-relevant content. These methods 
have been evaluated on benchmarks such as EgoSchema~\cite{mangalam2023egoschema} and Video-MME~\cite{fu2024videomme}. Despite these advances, existing long-context VLMs treat all content uniformly without structured modeling of user interactions or their inter-relations, and thus cannot capture the individual behavioral patterns necessary for genuine personalization.

\subsection{Egocentric Vision}
Egocentric data captured from wearable devices aligns with natural human perception 
and provides foundations for personal AI assistants~\cite{grauman2022ego4d, 
yang2025egolife}. The field has advanced significantly through large-scale datasets 
including EPIC-KITCHENS~\cite{damen2018scaling, damen2022rescaling}, 
Ego4D~\cite{grauman2022ego4d}, and EgoExo4D~\cite{grauman2024ego}, which cover a 
broad range of daily activities and human-object interactions from a first-person 
perspective. Building on these resources, multimodal pretraining has emerged as an 
effective approach for egocentric representation learning: EgoVLP~\cite{lin2022egocentric} 
pioneered this direction with an egocentric-aware contrastive loss over 3.8M clip-text 
pairs, EgoVLPv2~\cite{pramanick2023egovlpv2} incorporated cross-modal fusion via gating 
mechanisms, and LaViLa~\cite{zhao2023lavila} repurposes large language models as 
narrators to generate dense activity descriptions for contrastive learning.

Beyond action recognition, anticipating future actions is crucial for proactive 
assistance. The Anticipative Video Transformer~\cite{girdhar2021avt} established 
transformer-based attention over feature histories for action anticipation, with 
MAT~\cite{wang2023mat} and InAViT~\cite{roy2024inavit} further enhancing temporal 
and hand-object interaction modeling. EASG~\cite{rodin2024easg} introduces Egocentric 
Action Scene Graphs that capture temporally-evolving relationships between actions and 
objects, demonstrating effectiveness for both anticipation and activity summarization. 
Works such as EgoPlan~\cite{chen2023egoplan}, HD-Epic~\cite{perrett2025hd}, and 
EPIC-KITCHENS~\cite{damen2018scaling} provide detailed human-object interaction 
annotations for action and motion understanding, but focus on short-term behavioral 
segments without long-term user-specific behavioral organization, limiting their 
applicability to preference modeling and personalized assistance.

\subsection{Memory Systems}
Enabling large language models to maintain long-term, adaptive memory has attracted 
growing interest~\cite{zhong2024memorybank, packer2023memgpt}. MemoryBank~\cite{zhong2024memorybank} implements storage and retrieval with Ebbinghaus forgetting curve-inspired reinforcement for personality modeling, while MemGPT~\cite{packer2023memgpt} introduces hierarchical memory tiers where an LLM self-manages transitions between the main context and external archival storage. Benchmarks such as LongMemEval~\cite{wu2025longmemeval}, which assesses multi-session reasoning and temporal dynamics, and LoCoMo~\cite{maharana2024evaluating} reveals that even retrieval-augmented LLMs struggle with lengthy multi-session reasoning. Graph-structured representations have emerged as powerful tools for organizing relational knowledge: HippoRAG~\cite{gutierrez2024hipporag} combines knowledge graphs with Personalized PageRank for multi-hop reasoning, GraphRAG~\cite{edge2024graphrag} constructs entity-centric graphs through community detection and hierarchical summarization for query-focused retrieval, and Zep~\cite{rasmussen2025zep} introduces temporal knowledge graphs with hybrid retrieval combining semantic similarity, lexical matching, and structured graph traversal. Generative Agents~\cite{park2023generative} established a foundational architecture that stores complete experiences in natural language, synthesizes higher-level reflections, and retrieves relevant memories for planning, demonstrating emergent social behaviors in simulated environments.

In the egocentric domain, EgoLife~\cite{yang2025egolife} first highlighted the value of RAG-based memory for personalized suggestions; however, as user history grows, semantically similar yet contextually irrelevant entries increasingly overwhelm simple similarity matching, degrading retrieval precision at scale. AMEGO~\cite{goletto2024amego} proposes an active memory that dynamically organizes egocentric interactions into structured representations online. While these works have advanced memory mechanisms and explored egocentric applications separately, none simultaneously address the structured organization of long-term multimodal egocentric data and the internalization of user-specific behavioral patterns for genuine personalization. Our work bridges this gap by constructing a heterogeneous interaction graph with persistent entity nodes and compact user profiles, coupled with a self-supervised habit learning task that internalizes individual behavioral regularities beyond simple retrieval.

%% file: sections/3_method.tex
\section{Method}
\begin{figure*}[t]
\centering
\includegraphics[width=1\linewidth]{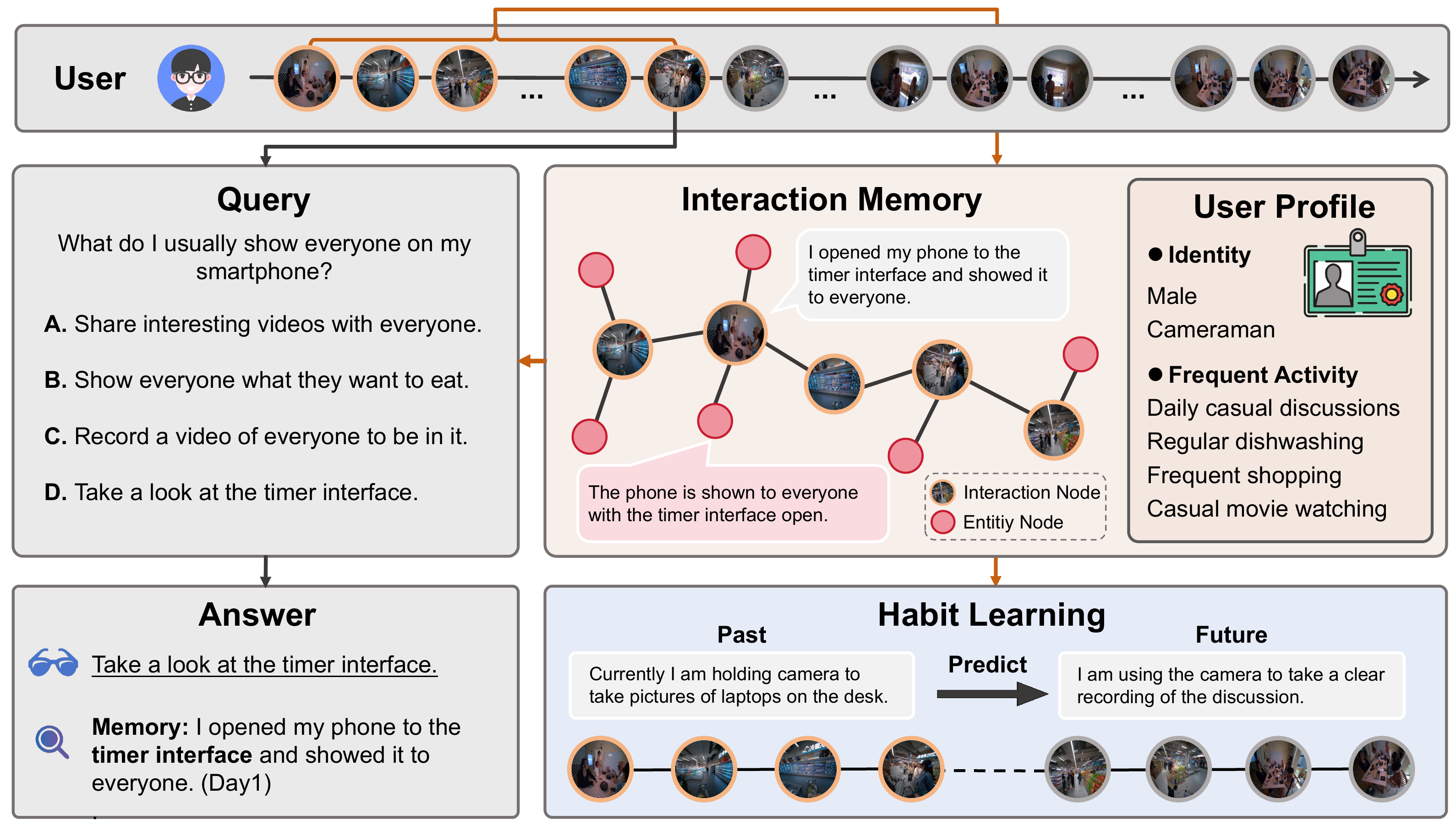}
\vspace{-15pt}
\caption{Framework overview of EgoSelf: a personalized egocentric assistant framework. EgoSelf constructs a heterogeneous graph-structured personal interaction memory, where nodes represent historical user interaction events, involved objects, and persons, while edges encode temporal, causal, and semantic relations among them. Based on this structured memory, the system extracts user-specific habit profiles that summarize long-term behavioral patterns, supporting personalized response generation and preference-aware recommendation. A dedicated habit learning task is designed to infer potential user habits by forecasting subsequent interactions from historical graph data. By integrating graph memory construction, habit profiling, and personalized prediction training, EgoSelf enables context-aware, behavior-consistent interactions for egocentric assistants.}
\vspace{-12pt}
\label{fig:intro}
\end{figure*}

In this section, we introduce EgoSelf, a framework for personalized egocentric assistants. As illustrated in Figure~\ref{fig:intro}, EgoSelf comprises several core components, a \textit{graph-based interaction memory} that encodes temporal, causal, and entity-level correlations from past egocentric observations, and an attached \textit{user profile module} that consolidates long-term behavioral patterns into a compact representation. Finally, we also include a \textit{habit learning task} that trains the model to predict future interactions from historical graph data, thereby capturing user-specific behavioral coherence.

\subsection{Problem Formulation}
We formalize the personalization task as follows. Given a user query $q_t$ arriving at time $t$, the assistant generates a response conditioned on both $q_t$ and the historical memory $\mathcal{M}_t$:
\begin{equation}
    \text{Ans} = \text{EgoSelf}(q_t, \mathcal{M}_t),
\end{equation}
where $\mathcal{M}_t = \{\mathcal{G}_t, \mathcal{P}\}$ denotes the memory state at time $t$, consisting of the interaction graph $\mathcal{G}_t$ constructed from all observations prior to $t$, and the user profile $\mathcal{P}$ summarizing long-term behavioral patterns. Crucially, the system may only access information available before query time, ensuring temporally valid inference. Our goal is to deliver accurate, personalized responses tailored to each user's individual history and preferences.

\subsection{Interaction Memory}
Motivated by cognitive theories of human memory organization~\cite{tulving1984relations}, we construct a heterogeneous graph $\mathcal{G} = (\mathcal{V}, \mathcal{E})$ as the memory backbone of EgoSelf. This structure mirrors the cognitive mechanism by which humans organize memories through both sequential context and shared semantic elements, aiming to unify fragmented experiences and enable consistent long-term personal user modeling.

\subsubsection{Node Definition.}
The node set comprises two categories:
\begin{equation}
    \mathcal{V} = \mathcal{V}_{\text{event}} \cup \mathcal{V}_{\text{entity}},
\end{equation}
where event nodes $\mathcal{V}_{\text{event}}$ capture episodic experiences from the user's daily life, and entity nodes $\mathcal{V}_{\text{entity}}$ encode persistent semantic knowledge about objects and people.

\paragraph{Event Nodes.}
We segment continuous egocentric video streams into discrete interaction events, each of which serves as a fundamental episodic unit in our memory graph. Rather than encoding all visual content, our memory focuses on meaningful interactions where the user actively engages with objects or people, as these moments carry the richest signals for understanding personal behavioral patterns. Formally, an event node $v_i \in \mathcal{V}_{\text{event}}$ is represented as:
\begin{equation}
v_i = (c_i, \mathcal{O}_i, \mathcal{S}_i, l_i, t_i),
\end{equation}
where $c_i$ is a language caption describing the wearer's activity, $\mathcal{O}_i$ denotes the set of interacted objects, $\mathcal{S}_i$ contains detected speech content, $l_i$ indicates the location, and $t_i$ is the timestamp. To obtain this structured representation, we feed each video segment along with its audio transcript into an MLLM, which extracts the activity description, involved objects, spoken content, and contextual information.

\paragraph{Entity Nodes.}
Entity nodes $\mathcal{V}_{\text{entity}}$ represent persistent semantic knowledge spanning disjoint events, enabling long-term tracking of objects and people across the user's history. By maintaining stable entity representations, we establish consistent anchors across disconnected events, avoiding fragmented memory and enabling reliable relation modeling. This design allows the system to track consistent user interactions over time. The edge set then $\mathcal{E}$ captures the associative structure of memory:
\begin{equation}
    \mathcal{E} = \mathcal{E}_{\text{event-event}} \cup \mathcal{E}_{\text{event-entity}}.
\end{equation}

\paragraph{Event-Event Edges.}
Inspired by the cognitive insight that humans understand their habits by linking related actions~\cite{bruner2009process}, we define three intuitive types of event-event connections.
\textit{Causal edges} link a prerequisite event to its consequent, identified by prompting an LLM to reason about explicit logical dependencies.
\textit{Co-activity edges} connect events belonging to the same high-level activity without strict causality.

\paragraph{Event-Entity Edges.}
We connect each event node to the object and person entities that participated in that event, thereby forming a connected subgraph. This structure supports entity-centric queries and enables the aggregation of interaction patterns around specific objects or people, which are essential for modeling user preferences over long-term history.

\paragraph{User Profile}
While the interaction graph captures fine-grained episodic details, we additionally maintain a user profile 
$\mathcal{P}$ to represent long-term stable traits and consistent habits in a compact form. Unlike fine-grained interaction records that focus on short-term events and specific contextual details, the user profile summarizes persistent user characteristics over an extended period, providing a holistic, condensed representation of user preferences. Such high-level and stable information is crucial for maintaining consistent behavior modeling over time, especially in long-term egocentric scenarios. The profile provides high-level contextual priors that complement the detailed graph memory during response generation.

To construct a meaningful profile, we first group discrete action records by their semantic meanings, clustering behaviors into coherent categories, and filter clusters by occurrence frequency, retaining high-frequency patterns that reflect stable habits while discarding sporadic behaviors. Ultimately, we abstract fine-grained behavioral details into high-level preference descriptions via LLM-based summarization. This process yields a structured user profile that provides stable contextual support for personalized reasoning and response generation. The profile remains as a complementary support document for each user and is sent to the retrieved context during inference.

\subsection{Memory Retrieval}
\label{sec:retrieval}
Given a query $q_t$, we retrieve relevant information from the graph memory $\mathcal{G}_t$ through a three-stage process. We first compute embedding similarities between the query and all nodes in $\mathcal{G}_t$.  It is also noted that all text embeddings are precomputed using Gemini Embedding~\cite{lee2025geminiembeddinggeneralizableembeddings} to enable efficient inference. Let $\text{sim}(q, v)$ denote the cosine similarity between the query embedding and the text embedding of node $v$. Here, We select top-$k$ entity nodes and top-$k$ event nodes based on similarity scores:
\begin{align}
\mathcal{V}^{\text{cand}}_{\text{entity}}
&= \text{TopK}_{v \in \mathcal{V}_{\text{entity}}}(\text{sim}(q, v), k), \\
\mathcal{V}^{\text{cand}}_{\text{event}}
&= \text{TopK}_{v \in \mathcal{V}_{\text{event}}}(\text{sim}(q, v), k).
\end{align}

\noindent \textbf{Graph-Constrained Filtering.}
To ensure retrieval consistency and reduce irrelevant noise, we retain only event nodes that are connected to at least one selected entity node in the graph memory:
\begin{equation}
\mathcal{R} = \{v \in \mathcal{V}^{\text{cand}}_{\text{event}} \mid \exists\, e \in \mathcal{V}^{\text{cand}}_{\text{entity}},\, (v, e) \in \mathcal{E}_{\text{event-entity}}\}.
\end{equation}
This intersection operation filters out potential incorrect matches that lack entity-level grounding, effectively narrowing down the candidate event set to those with strong meaningful semantic associations to the query-relevant entities. This step further enhances the precision of retrieval by ensuring all retained events have explicit relational ties to the entities identified as highly relevant to the user's query.

\noindent \textbf{Context Expansion.}
Finally, building on the filtered event set from the prior entity-grounding step, we expand the core set $\mathcal{R}$ by incorporating multi-hop neighboring events to provide comprehensive context for downstream reasoning:
\begin{equation}
\mathcal{R}^{+} = \mathcal{R} \cup \{v \in \mathcal{V}_{\text{event}} \mid \exists\, u \in \mathcal{R},\, (u, v) \in \mathcal{E}_{\text{event-event}}\}.
\end{equation}
This filter-then-expand design ensures that we only enrich context based on highly relevant and well-grounded events, avoiding the introduction of noisy or unrelated information during expansion. The initial filtering guarantees retrieval reliability, while the later expansion complements the necessary temporal and relational context for a complete understanding. The final retrieved events $\mathcal{R}^{+}$, together with the user profile $\mathcal{P}$, form the complete structured context for subsequent response generation.

\begin{figure}[t]\centering
\includegraphics[width=0.82\linewidth]{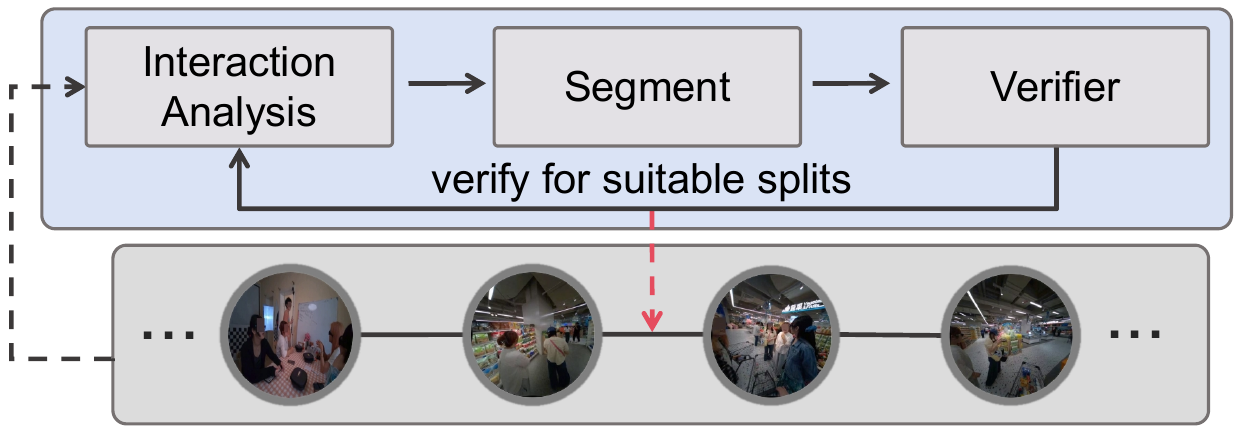}
\vspace{-2pt}
\caption{Illustration of habit learning task generation. We identify a suitable partition point in the interaction graph using a reasoning LLM that leverages event relationships, and then split the video into past and future segments to form training pairs. An additional LLM verifier examines the generated pairs and refines captions to ensure reliable training data.}
\vspace{-18pt}
\label{fig:dataset}
\end{figure}

\subsection{Habit Learning}
\label{sec:habit_learning}
To enhance the model's ability to capture user-specific behavioral patterns, we introduce a self-supervised habit-learning task that leverages the graph structure. Unlike generic behavior modeling objectives that focus on shared patterns across users, this task explicitly targets the stable and recurring routines that define individual behavior, which is crucial for achieving personalized and robust behavioral modeling.

Motivated by behavioral psychology research~\cite{bruner2009process, watson1913psychology, skinner2019behavior, lally2010habits}, habits are formed through interconnected and sequential activities and can be inferred from observable activity patterns. Specifically, human behavioral habits are not random or isolated events but arise from repeated temporal dependencies, contextual consistencies, and structural correlations between consecutive actions. Embedding these psychological insights into a self-supervised learning paradigm allows the model to discover intrinsic behavioral regularities without manual annotations or domain-specific heuristics. The graph structure provides a natural representation for modeling relational dependencies between entities, actions, and environments, making it well-suited for encoding the interconnected nature of habitual behavior.

\noindent \textbf{Task Formulation.}
We formulate habit learning as a self-supervised learning task. Given a chain of event nodes $\mathcal{V}_{\text{event}} = \{v_1, v_2, \ldots, v_n\}$ ordered by timestamp, we select a partition point $t$ and split the sequence into observed history $\mathcal{V}_{\leq t} = \{v_1, \ldots, v_t\}$ and future events $\mathcal{V}_{> t} = \{v_{t+1}, \ldots, v_n\}$. The model is trained to predict a summary of $\mathcal{V}_{> t}$ conditioned on the video frames and graph structure corresponding to $\mathcal{V}_{\leq t}$.

This formulation encourages the model to perform user-centric reasoning: given a specific user's historical interactions, what personalized interactions might they perform next? Through this prediction objective, the model implicitly learns to capture individual behavioral coherence and user-specific causal associations.

\noindent \textbf{Data Generation.}
As illustrated in Figure~\ref{fig:dataset}, we construct training pairs through a three-stage pipeline. A reasoning LLM (DeepSeek-R1~\cite{guo2025deepseek}) will analyze the event graph to identify semantically coherent interaction scenes and determine suitable partition points based on causal and temporal relationships between events. We then partition the video into past and future segments at the identified partition points, forming (history, future) training pairs.

Finally, an additional LLM verifier examines the generated pairs for coherence and refines the captions to ensure reliable training data. This personalized prediction objective provides a tailored, user-specific self-supervised signal. By predicting a user's unique upcoming interactions, the model learns behavioral patterns that generalize to downstream personalization tasks.

%% file: sections/4_experiments.tex
\section{Experiments}
We evaluate EgoSelf on the EgoLife dataset~\cite{yang2025egolife}, a benchmark for personalized egocentric research. EgoLife collects multimodal daily activities of six participants over seven days in a controlled environment, where user-specific behaviors during Earth Day party preparations are recorded. The dataset captures distinct individual behavioral preferences and interaction habits, with all Chinese annotations translated into English. It includes five specific types of questions for a single participant, namely EntityLog, EventRecall, HabitInsight, RelationMap, and TaskMaster, which target fine-grained understanding of personal habits, daily patterns, and contextual interaction characteristics to provide a comprehensive evaluation of personalized egocentric modeling.

\begin{table*}[h]
\centering
\caption{Performance comparison on the EgoLifeQA benchmark.}
\label{tab:model_egolifeQA}
\resizebox{1.0\textwidth}{!}{
\begin{tabular}{l|ccccc|c}
\toprule
\textbf{Model} & \textbf{EntityLog} & \textbf{EventRecall} & \textbf{HabitInsight} & \textbf{RelationMap} &
\textbf{TaskMaster} & \textbf{Average}\\ \midrule
\textbf{Gemini}~\cite{team2023gemini}       & 36.0 & 37.3 & 45.9 & 30.4 & 34.9 & 36.9 \\
\textbf{GPT-4o}~\cite{hurst2024gpt}         & 34.4 & 42.1 & 29.5 & 30.4 & 44.4  & 36.2\\
\midrule
\textbf{LLaVA-OneVision}~\cite{li2024llava} & 36.8 & 34.9 & 31.1 & 22.4 & 28.6  & 30.8 \\
\textbf{EgoGPT}~\cite{yang2025egolife}      & 39.2 & 36.5 & 31.1 & 33.6 & 39.7 & 36.0 \\
\textbf{EgoSelf (Ours)}                     & 38.4 & 42.9 & 44.3 & 36.0 & 41.3 & 40.6\\
\bottomrule
\end{tabular}
}
\end{table*}

\subsection{Experimental Setup}
We extract discrete interaction nodes from 30-second egocentric video clips to construct the interaction graph. 
Whisper~\cite{radford2023whisper} transcribes audio from video streams. Sentence Transformer~\cite{reimers2019sentencebert} computes text similarity for preliminary object and entity matching.
Grounding DINO~\cite{liu2023groundingdino} verifies visual consistency by computing IoU between detected and tracked bounding boxes.
Qwen~\cite{bai2023qwen}, GPT-4o~\cite{openai2024gpt}, and Gemini-1.5-Flash~\cite{gemini2023technical} serve as core multimodal models for parsing video and audio inputs, extracting structured event representations, and inferring relational connections among events and entities.
DeepSeek-R1~\cite{deepseek2024r1} and GPT-4o generate training data for the habit learning task by mining recurring behavioral patterns from the constructed interaction graph.
Gemini embeddings~\cite{gemini2023technical} encode text captions for efficient similarity-based retrieval.

\subsection{Results}
Table~\ref{tab:model_egolifeQA} summarizes the performance of evaluated methods on the EgoLifeQA benchmark. Our approach, EgoSelf, attains state-of-the-art average performance across all benchmarks, surpassing existing strong baselines and specialized egocentric models by a clear margin. It also exhibits consistent and clear performance advantages over representative other baseline models across most task categories in Egolife Dataset.

EgoSelf delivers superior performance on tasks involving temporal event recall and relational interaction modeling, demonstrating strong capabilities in temporal understanding and social relationship modeling. On tasks related to behavioral habit understanding, it achieves highly competitive performance, validating that the proposed graph-based memory mechanism effectively encodes and reasons about habitual patterns from egocentric data. The consistent gains further highlight the benefits of the entity-centric graph structure for person-centric identification and interaction analysis.

\begin{table}[t]
\centering
\caption{Performance comparison across different history lengths. Longer history spans require reasoning over more temporally distant events and accumulated behavioral data.}\label{tab:history_length}
\vspace{-6pt}
\begin{tabular}{lcc}
\toprule
\multirow{2}{*}{\textbf{Method}} & \multicolumn{2}{c}{\textbf{History Length}} \\
& $<$2h & $>$24h \\
\midrule
\textbf{EgoGPT}~\cite{yang2025egolife} & 28.2  & 25.0 \\
\textbf{EgoGPT+EgoRAG}~\cite{yang2025egolife} & 27.2  & 35.4 \\
\midrule
\textbf{EgoSelf } & 31.3  & 38.8\\
\bottomrule
\end{tabular}
\vspace{-12pt}
\end{table}
\subsection{Effectiveness on Different History Length}
We analyze how performance varies with the temporal span of questions. Table~\ref{tab:history_length} groups questions by the history length required to answer them. From the experimental results, EgoSelf maintains stable, consistent performance across both short- and long-range queries, including those that demand reasoning over user histories longer than $24$ hours. The user profile integrated into EgoSelf captures stable behavioral traits, providing rich and coherent context for long-term reasoning. For shorter temporal spans, the framework leverages fine-grained graph structure and precise retrieval to generate more accurate, context-matched responses. In contrast,  other methods exhibit clear performance degradation on long-range queries, as they rely on fragmented retrieval without structured memory or persistent user profiling.

\subsection{Effectiveness of Retrieval}
\begin{figure*}[t]
\centering
\includegraphics[width=0.95\linewidth]{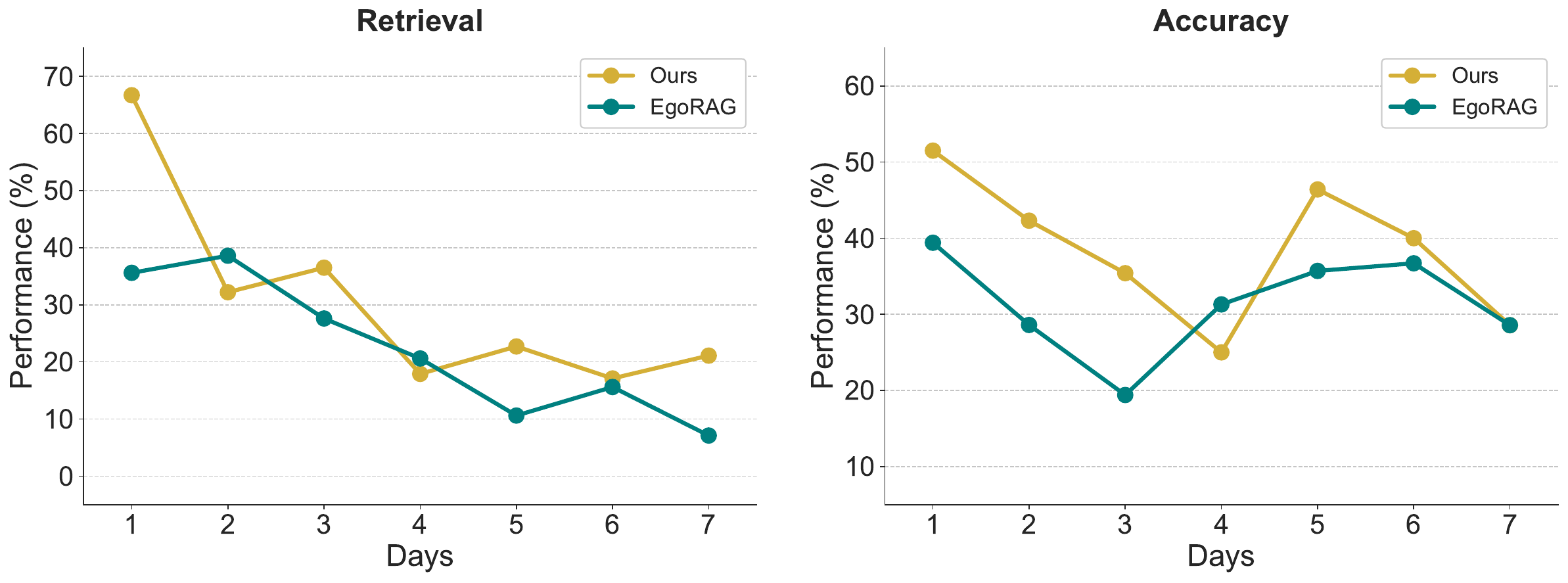}
\vspace{-6pt}
\caption{The retrieval and accuracy performance comparison between EgoSelf and EgoRag on the 7‑day task. Yellow line represents our EgoSelf Performance, and green line represents the compared baseline.}
\label{fig:comparison_ego_rag}
\vspace{-12pt}
\end{figure*}

As shown in Figure~\ref{fig:comparison_ego_rag}, our EgoSelf method achieves a significantly higher hit rate than EgoRag across the 7-day egocentric task with metric Hit@$k$ where $k$ is set to $7$, which evaluates whether the model can retrieve relevant candidates within a predefined temporal window around the ground-truth target event. In addition, EgoSelf maintains relatively stable and robust performance throughout the entire evaluation period with only minor performance drops. In comparison, EgoRag exhibits a clear and continuous performance degradation over time, resulting in a much lower overall score. These results clearly demonstrate that the structured memory design of EgoSelf effectively maintains reliable and consistent retrieval in long-term egocentric scenarios, as it enables more accurate modeling and persistent retention of long-term user preferences and behavioral patterns, allowing the system to locate highly relevant information even as the temporal context extends over multiple days, thus avoiding severe performance decay observed in the baseline method.

Moreover, We further evaluate model performance across progressively extended temporal contexts, ranging from one to seven days correspond to the retrival performance. The proposed method consistently maintains stable and superior performance across all evaluated temporal durations, demonstrating strong robustness to varying query lengths and memory horizons. As the temporal window expands, our approach continues to effectively capture and reason over accumulated visual and semantic information, without degrading the quality of understanding or retrieval. In contrast, baseline methods exhibit clear performance fluctuations and progressive drops when faced with longer temporal spans and extended query contexts. These results validate that our framework can reliably handle long-range temporal dependencies and maintain effective reasoning across extended egocentric observation periods.

\begin{figure*}[t]
\centering
\includegraphics[width=1.0\linewidth]{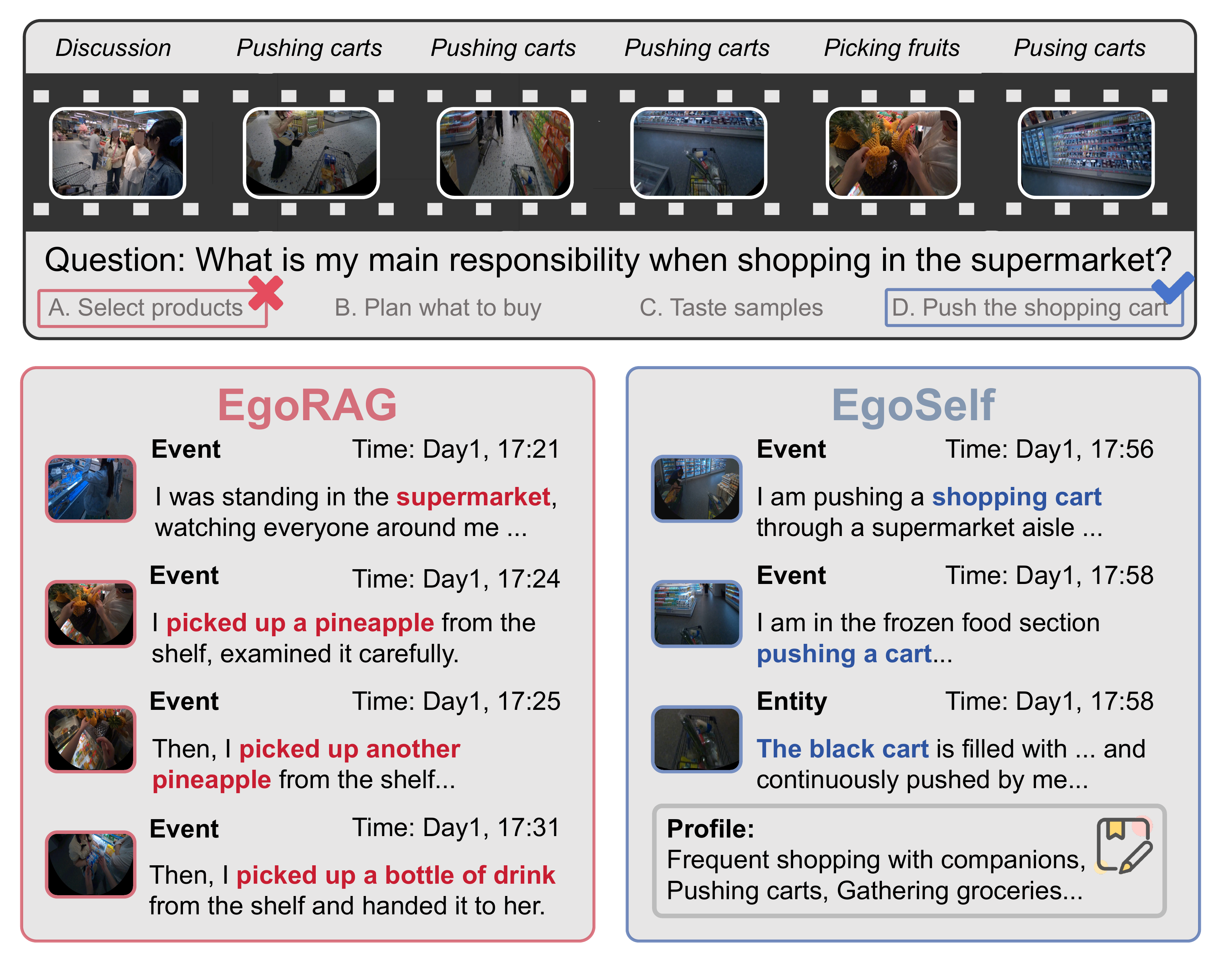}
\vspace{-18pt}
\caption{Qualitative comparison between EgoSelf and EgoRAG. EgoSelf retrieves structured, relationally connected events that capture consistent behavioral patterns, with the user profile supplementing long-term habit context. In contrast, EgoRAG retrieves isolated events without relational modeling, leading to a fragmented and less reliable reasoning context.}
\label{fig:comparison_fig}
\vspace{-12pt}
\end{figure*}
\subsection{Qualitative Analysis}
Figure~\ref{fig:comparison_fig} compares EgoSelf with EgoRAG~\cite{yang2025egolife} on a representative example. EgoSelf retrieves relevant interaction records, organized into structured events, and aggregates long-term habits across events into the user profile. During retrieval, the graph structure enables the extraction of causally and temporally related events, grounding responses in consistent behavioral patterns.
This structured retrieval also filters out irrelevant interactions, ensuring that the reasoning context remains focused and semantically consistent.
The user profile further supplements temporal and causal relations with persistent behavioral priors, enhancing the consistency of the final output. In this case, using our graph intersection retrieval, we filter out most noisy context and retain key contextual information, while the user profile also provides complementary personal context for personalization.

In contrast, baseline method retrieves isolated activity events without relational modeling. Lacking structured memory organization, it fails to capture consistent user habits or inferential connections across interactions, leading to context-inconsistent outputs. In the example shown, EgoSelf successfully extracts several target events, along with structurally connected related events. EgoRAG fail to retriece most important activities; although it occasionally identifies the certain relevant target behavior, the lack of relational filtering introduces noise, leading to incorrect predictions. Thus, it includes many instances, such as selecting goods activities, which confuse the model and cause it to output false results.

This comparison illustrates how EgoSelf's graph-based memory effectively integrates discrete episodic interactions into a coherent, habit-aware structure, thereby supporting more consistent, accurate, and personalized responses tailored to individual user behaviors. This qualitative analysis further validates the efficacy of explicitly modeling relational connections and fine-grained user profiles, rather than relying solely on unstructured content retrieval that lacks contextual and behavioral awareness. By constructing robust, continuous behavioral chains that link related events over time, EgoSelf successfully maintains contextual consistency across long-term interaction histories, which is a critical and indispensable factor for achieving reliable and truly personalized modeling in practical egocentric assistant systems.

%% file: sections/5_conclusion.tex
\section{Conclusion}
In this work, we introduce EgoSelf, a framework for personalized egocentric assistants. EgoSelf builds a graph-based memory to store historical user interactions, where related people and objects are represented as individual nodes. Relations between events, entities, and timestamps are encoded as edges, forming a structured and interpretable knowledge base. A user profile is also created to summarize personal habits, preferences, and long-term behavioral patterns, supporting personalized responses and recommendations. Based on cognitive insights, we design a learning task that uses action prediction to help the model capture user habits. This task extracts relevant information from the graph and trains the model to predict future events using historical data. With personalized training, user profiling, and graph retrieval, EgoSelf supports context-aware interactions that align closely with user behavior patterns, thereby enabling natural and adaptive long-term human-AI collaboration.